\documentclass[10pt,twocolumn,letterpaper]{article}

\usepackage{cvpr}              
\usepackage{amssymb} 
\usepackage{amsmath}
\usepackage{graphicx} 
\usepackage{booktabs}
\usepackage{multirow}
\usepackage{array}
\usepackage{makecell}
\usepackage{algorithm}
\usepackage{algpseudocode}
\usepackage{xcolor}

\definecolor{cvprblue}{rgb}{0.21,0.49,0.74}
\usepackage[pagebackref,breaklinks,colorlinks,allcolors=cvprblue]{hyperref}

\def\paperID{12040} 
\def\confName{CVPR}
\def\confYear{2026}

\title{Property-Informed Diffusion-Based Text-to-Microstructure Generation}

\author{
	Bingxuan Dai\textsuperscript{1 \textdagger}, \; Hongsong Wang\textsuperscript{2,3 \textdagger}\thanks{Corresponding Authors. \textsuperscript{\textdagger}{Equal Contribution}}, \; Jie Gui\textsuperscript{1,4,5 *}\\
    $^{1}$School of Cyber Science and Engineering, Southeast University, Nanjing 210096, China\\
	$^{2}$School of Computer Science and Engineering, Southeast University, Nanjing 210096, China \\
	$^{3}$Key Laboratory of New Generation Artificial Intelligence Technology and Its Interdisciplinary \\
	Applications (Southeast University), Ministry of Education, China \\ 
	$^{4}$Purple Mountain Laboratories, Nanjing 210000, China \\
	$^{5}$Engineering Research Center of Blockchain Application, Supervision And Management\\ (Southeast University), Ministry of Education, China \\
	\tt\small\{220245799, hongsongwang, guijie\}@seu.edu.cn \\
}

\begin{document}
\maketitle

\begin{abstract}
Designing 3D metamaterial microstructures that meet the intended functions remains a major challenge, as it typically requires domain expertise, iterative simulations, and extensive manual tuning. Existing work on inverse design that automatically generates microstructures based on desired target properties often suffers from limited design diversity and faces challenges in ensuring the physical feasibility of the generated structures. To address this issue, a property-informed diffusion-based network is proposed that enables the generation of 3D microstructures directly from textual descriptions. Unlike traditional property conditioning methods, our approach leverages rich guidance in terms of semantics and physical properties in the text input to support diverse structure synthesis. To enforce consistency between the generated structures and the target textual prompts, a dual alignment strategy is adopted, including contrastive text-structure alignment and test-time reward-guided alignment. Experimental results show that the model is capable of generating semantically meaningful and physically plausible structures across a wide range of material categories. Our approach has good potential for interactive microstructure design and opens up new directions for combining language-based interfaces with inverse material discovery. Code is available at: \url{https://github.com/hongsong-wang/PropDiff-TMG}.
\end{abstract}

\section{Introduction}
Emerging materials~\citep{chen2021simultaneously, xia2021review, divilov2024disordered,gao2026generative}, characterized by distinctive microstructures, enhanced performance, and novel functionalities, have garnered increasing attention due to the rapid advancement of scientific research and engineering innovation. Their integration into high-technology sectors such as aerospace \citep{mcmillan2022inverse}, environmental protection \citep{mohsenizadeh2018additively}, and biomedical engineering \citep{suhas2025review} has substantially expanded both the functionality and application potential of materials. The properties of these materials are predominantly governed by their internal microstructures rather than their constituent substances. Consequently, the design and engineering of microstructures have assumed a central role in the development of novel materials, enabling the optimization of performance and the customized realization of specific functionalities.

\begin{figure}[t]
    \centering
    \includegraphics[width=0.5\textwidth]{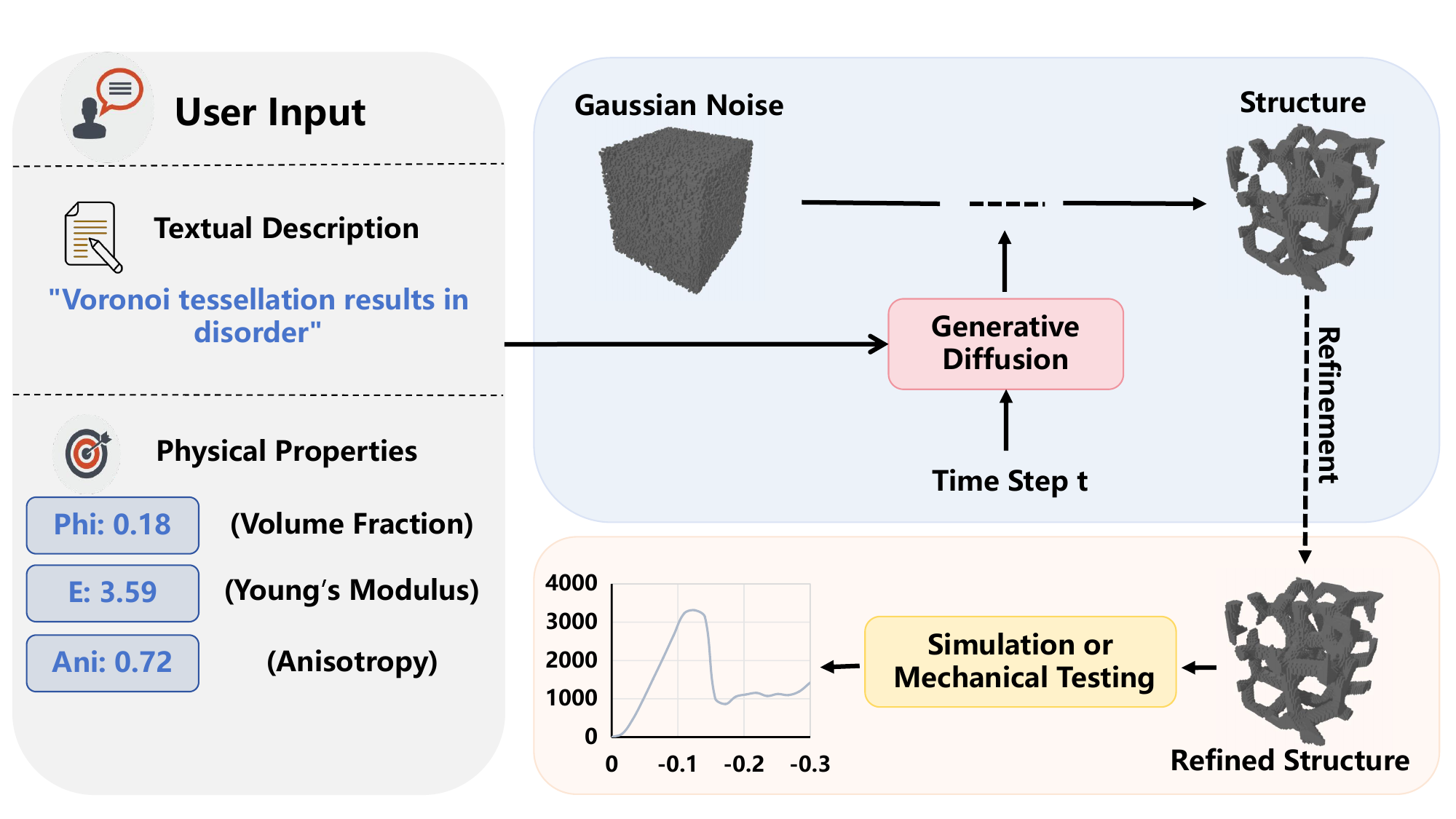}
    \vspace{-0.2cm}
    \caption{\textbf{Overview of our approach:} A diffusion-based framework can generate metamaterial microstructures according to the textual description and physical properties.}
    \label{fig:overview}
    \vspace{-0.3cm}
\end{figure}

Inverse design has emerged as a powerful paradigm in materials science, shifting the traditional trial-and-error process toward a goal-driven approach~\cite{zheng2023deep,wang2022inverse,gao2025polymer,gao2026generative}. In mechanical metamaterials research, phase-field modeling~\citep{zhao2023development} and topology optimization~\citep{sha2024topology, wu2021topology} have been used to inversely design architectures with extraordinary characteristics such as extreme stiffness or auxetic behavior~\cite{han2024inverse}. Meanwhile, data-driven and machine learning techniques have accelerated inverse design by enabling rapid exploration of vast microstructural design spaces~\cite{kollmann2020deep,lee2023machine,gao2026generative}.

Given the specific physical properties of the microstructure, experienced material scientists can inversely design the corresponding microstructures using computational approaches such as phase-field modeling~\citep{kumar2020inverse}, mathematical modeling~\citep{zheng2021mathematically}, and topology optimization~\citep{zhou2022computational}. Although these traditional methods are highly interpretable and follow physical principles, they often rely on extensive domain expertise, hand-crafted design spaces, and computationally expensive iterative solvers. Although recent advances in deep learning have made data-driven material structure design possible, most existing methods \citep{choudhary2022recent, zheng2023deep, seo2023dl} still rely on expert-defined conditions or parameter controls. In contrast, the ability to generate metamaterial microstructures directly from textual descriptions remains underexplored, limiting the accessibility of such models. 

To address these problems, we propose a property-informed diffusion-based text-to-microstructure generation (PropDiff-TMG), a fully automated, robust, and generalizable framework for the generation of metamaterial microstructures. As shown in Figure~\ref{fig:overview}, our method accepts text as input and generates high-quality microstructures. 

To achieve stable and targeted microstructure generation based on structural text feature descriptions, we adopt a self-conditional diffusion framework to gradually guide the generation results to approach the semantic target during the denoising process. On this basis, we further introduce the random injection of physical properties to modulate the generation process to improve the consistency and controllability of the structure in terms of mechanical properties. Specifically, we adopt the feature-wise linear modulation (FiLM)~\citep{perez2018film} to jointly encode physical and semantic information into a dynamic adjustment signal at the feature level, thereby achieving finer-grained generation control. After training, the model can generate diverse and physically consistent 3D microstructures based on material properties and structural text descriptions.

To enhance the semantic and physical consistency of the generated structures, a dual alignment optimization strategy is proposed. During training, a contrastive text-structure alignment is introduced to align the text encoder with the visual encoder through contrastive learning to alleviate the semantic shift of the general language model~\citep{radford2018improving, brown2020language} in the material domain. During testing, a reward-guided alignment strategy is employed to enhance the synergy between semantics and physics by constructing a composite reward function that combines CLIP similarity and discriminator authenticity, and performing normalization. To further improve local fidelity, we introduce a multi-round local editing mechanism, where candidate regions are sampled and refined based on reward signals in each iteration, making high-scoring regions easier to retain and enhance, thereby optimizing the physical rationality of the structure while maintaining global consistency.

Together, these components enable our framework to generate high-quality, diverse, and physically meaningful 3D microstructures from natural language descriptions. To train and evaluate the model, we conduct experiments on a dataset of paired text descriptions and voxelized microstructures comprising multiple types of materials. Both quantitative evaluations and simulation experiments show that our approach achieves promising results in terms of both semantic alignment and physical plausibility. In summary, our contributions are as follows:
\begin{itemize}
\item \textbf{Innovative inverse design paradigm:} We introduce a self-conditional diffusion framework for generating metamaterial microstructures from textual descriptions and physical properties.
\item\textbf{Effective cross-modal alignment mechanism:} We propose a dual alignment mechanism integrating cross-modal contrastive training and reward-guided optimization to enforce semantic and physical consistency.
\end{itemize}

\section{Related Work}

\noindent\textbf{Inverse Design of Microstructure:}
Traditional methods~\citep{sigmund2009systematic, wang2021design, zhang2023mechanics, gao2020comprehensive} rely on established physical principles and empirical trial-and-error processes to design 3D structures, but often suffer from long design cycles, high computational costs, and considerable algorithmic complexity~\citep{du2018connecting, garner2019compatibility}.
To address these limitations, data-driven microstructure generation~\citep{jiao2021artificial, song2024artificial, lee2024data}, which is based on generative artificial intelligence, has emerged. For example, inverse design of metamaterials based on diffusion models~\citep{yang2024guided, bastek2023inverse} enables end-to-end generation conditioned on mechanical properties.
However, these approaches typically depend on domain-specific priors and require expert-driven condition design.

 While substantial progress has been made in text-guided 3D object generation~\citep{vahdat2022lion, zhao2023michelangelo, wei2023taps3d}, the design of metamaterial microstructures poses greater challenges due to their intricate physical, mechanical, and functional characteristics. Current works~\citep{yang2021words, hsu2022generative, buehler2022generating} typically employ textual conditions to generate 2D representations of material structures, which are subsequently transformed into 3D structures via post-processing, rather than an end-to-end generative framework. 
Txt2Microstruct-Net~\citep{zheng2024text} leverages a variational autoencoder (VAE) to generate 3D voxels and a CLIP-based module to align textual prompt with 3D voxels in the latent space. However, aligning textual prompts with 3D voxels using simple multi-layer perceptrons is challenging, and this approach requires multi-stage training. ChatMetamaterials~\cite{gao2026generative} enables efficient exploration of mechanical metamaterials by introducing a large language model–driven design engine.
 
\begin{figure*}[t]
    \centering
    \includegraphics[width=\textwidth]{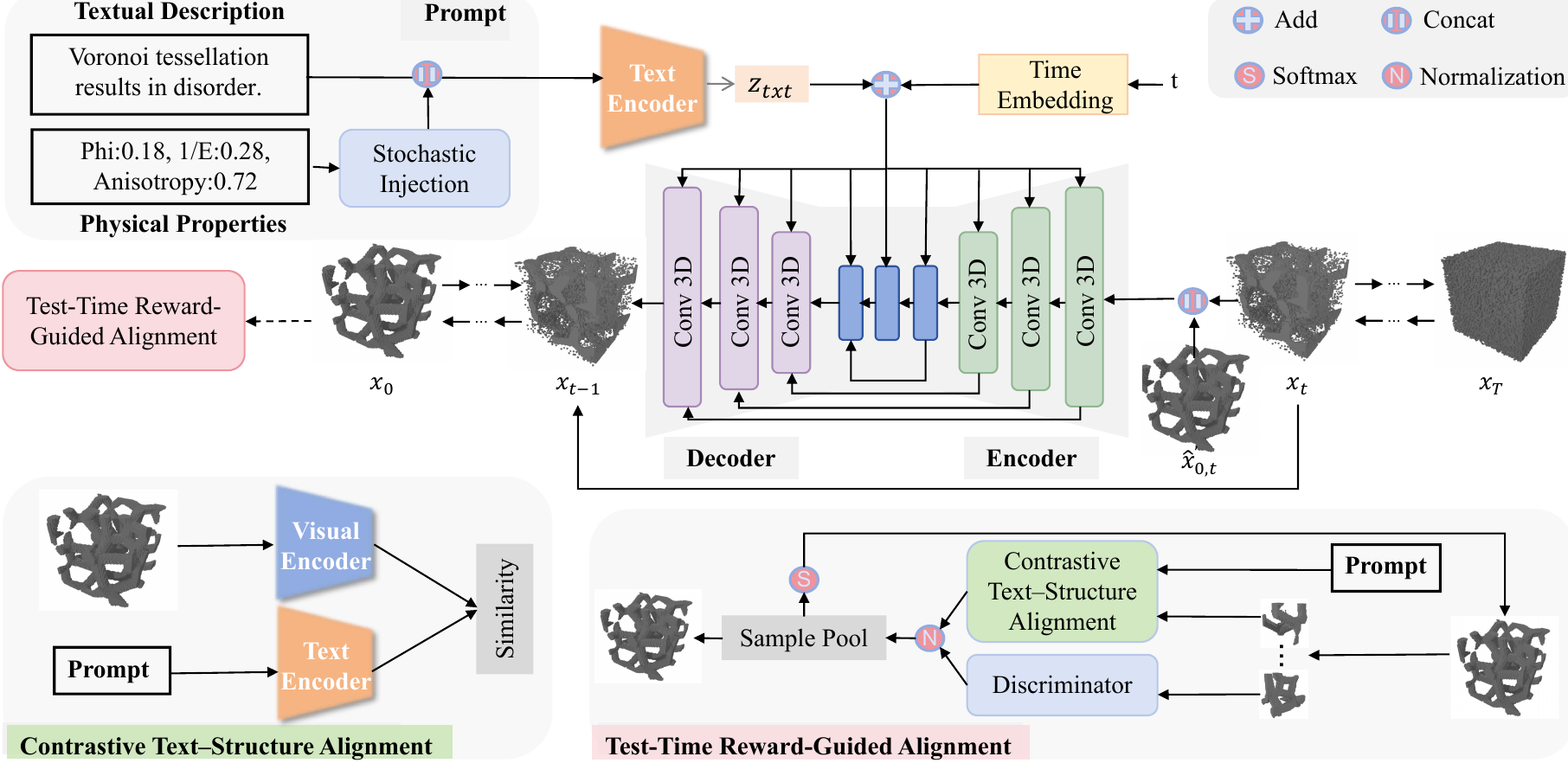}
    \caption{\textbf{Pipeline of the proposed PropDiff-TMG:} Our method is based on a self-conditional diffusion model, guided by textual descriptions and optionally injected physical properties. During training, a contrastive text–structure alignment strategy aligns text and structure representations via a contrastive loss. At inference, we further refine generations via reward-guided sampling using a CLIP and a discriminator, achieving dual alignment from both pretraining and optimization perspectives.}
    \label{fig:pipeline}
\end{figure*}

\noindent\textbf{Diffusion-Based Material Generation:}
Recently, diffusion models have made breakthrough progress in multiple generation tasks~\citep{wang2025diffusion, gao2024cat3d, zheng2025diffumeta,anciukevivcius2023renderdiffusion}, significantly improving the quality and controllability of material generation. 
The condition-guided joint diffusion framework can effectively generate materials with complex structures and high fidelity. TGDMat \citep{das2025periodic} is proposed as a text-guided joint diffusion model to achieve synchronous generation of atomic types, coordinates, and lattice structures in periodic materials. On the other hand, in response to the problems of multi-view consistency and illumination changes, MaterialMVP \citep{he2025materialmvp} effectively generates multi-view physical rendering textures with invariant illumination through the attention mechanism and consistency regularization. Material Anything \citep{huang2025material} generates physically based materials on any 3D object based on an end-to-end diffusion network. 
Diffusion-based methods provide an efficient and versatile framework for material generation, enabling integration of multimodal conditions like text, geometry, and physical properties beyond traditional approaches. Our work focuses on diffusion-based microstructure generation, aiming to optimize material performance through structural design.

\noindent\textbf{Physically-Based Material Generation:}
Enforcing physical constraints in 3D generation ensures material properties follow real-world principles. 3DTopia-XL \citep{chen20253dtopia} generates high-resolution 3D geometry and Physically-Based Rendering (PBR) materials via diffusion for property optimization. Diffusion Renderer \citep{liang2025diffusion} ensures consistent lighting by jointly estimating material and illumination. MatFuse \citep{vecchio2024matfuse} and Alchemist \citep{sharma2024alchemist} enable precise control of material properties under varying viewpoints and lighting. RichDreamer \citep{qiu2024richdreamer} enhances geometric detail and lighting consistency by jointly generating normals and depth maps. MaterialMVP \citep{he2025materialmvp} and SViM3D \citep{engelhardt2025svim3d} improve multi-view consistency and lighting invariance. Inspired by recent material generation research, we incorporate physical property constraints into text prompts to ensure structural integrity and functionality.

\section{Approach}
PropDiff-TMG is a unified diffusion-based framework for the inverse design of 3D metamaterial microstructures conditioned on semantic text and physical properties, as shown in Figure~\ref{fig:pipeline}. The entire design process is divided into three key stages to ensure controllability, generalization, and fidelity. First, we adopt a self-conditional diffusion model to generate microstructures from semantic descriptions. Second, we propose a random conditional injection mechanism based on physical knowledge, in which physical property descriptors are incorporated as auxiliary text prompts to refine the generative process for better consistency with specified physical properties. 
Third, we introduce a dual alignment strategy to enforce semantic and structural consistency further. This includes: (1) a CLIP-based contrastive encoder to align textual and structural representations during pre-training; and (2) a reward-guided sampling module that optimizes the diffusion process using a learned evaluator based on semantic and geometric alignment scores.

\subsection{Diffusion-Based Text-to-Microstructure}
This section introduces a diffusion-based inverse design framework that constitutes the foundation of our methodology. The framework is designed to generate 3D microstructures that align with high-level semantic descriptions while maintaining structural plausibility, thereby enabling precise and controllable mapping between structural geometry and functional performance, where each 3D microstructure is represented as a voxel grid capturing material occupancy in space.
Initially, the model is based on the denoising diffusion probabilistic model (DDPM) \citep{ho2020denoising}, where the forward process Gaussian denoising scheme, gradually perturbs the clean samples \(x_0\) into noisy samples \(x_t\), defined as:
\begin{equation}
    x_t=\sqrt{\gamma_t}\ x_0+\sqrt{1-\gamma_t}\epsilon,
\end{equation}
where \(\epsilon \sim \mathcal{N}(0, I)\) and \(\gamma_t\) decreases steadily from 1 to 0. In the reverse process, a U-Net-like 3D network is employed to reconstruct \(x_0\) from the noisy input \(x_t\).

To improve reconstruction quality and stability, a self-conditional diffusion model \citep{chen2022analog} is adopted, which recursively incorporates the previous prediction of the model as an auxiliary input \(\hat{x}_0'\) at each denoising step. By leveraging contextual information beyond the current noisy input, the model iteratively refines its estimations to achieve more accurate reconstruction. To mitigate excessive dependence on prior outputs and reduce cumulative errors, the auxiliary input is randomly replaced with zero during training with a fixed probability, thereby improving robustness.
Specifically, the network outputs:
\begin{equation}
    \hat{x}_0 = f(x_t, \hat{x}_0', t, z_i^{d}),
\end{equation}
where \(\hat{x}_0'=f(x_t, 0, t, z_i^{d})\) is either the previous prediction of the model (with 50\% probability) or zero. The text condition \(z_i^{d}\) is encoded using a pretrained text encoder, followed by sinusoidal time embeddings, and integrated into the denoising network via classifier-free guidance. The model is optimized by minimizing the reconstruction loss:
\begin{equation}
    \mathcal{L}_{con} = E_{\epsilon, t} \left\| \hat{x}_0 - x_0 \right\|_2^2,
\end{equation}
which encourages the alignment of text semantics with geometric structures through iterative denoising.

\subsection{Property-Informed Stochastic Conditioning}
To effectively incorporate physical property guidance into the inverse design process, we represent quantitative material physical property as augmented textual conditions, enabling seamless integration with semantic prompts. These enriched conditions are embedded and injected into the diffusion model using a FiLM-based modulation mechanism, which adaptively modulates intermediate network features to robustly guide microstructure generation toward satisfying both semantic and physical constraints.

\noindent\textbf{Physical Property Text Encoding:}
To enable inverse design optimized by quantitative physical properties, we augment the textual description \(T\) with explicit material properties such as Young’s modulus \(E\), isotropy index \(I\), and volume fraction \(V_f\), represented in descriptive textual form. The combined conditioning \(\tilde{T} = \{T, E, I, V_f\}\) guides the diffusion model \(\mathcal{G}\) to generate microstructures \(x\) that not only adhere to semantic constraints but are also optimized to meet the specified physical targets:
\(
    x \sim \mathcal{G}(\tilde{T}).
\)
we apply random masking to the physical property conditioning \(P = \{E, I, V_f\}\) during training. Each property \(p \in \{E, I, V_f\}\) is retained independently with probability \(r_p\), enabling the model to learn from both fully and partially specified prompts. This stochastic injection strategy allows the model to flexibly handle various conditioning scenarios, both with and without property constraints, and better align generation with physical targets.
To assess the physical fidelity of generated structures, we train a regression network \(\mathcal{R}\) using paired data \(\{(x_i, P_i)\}_{i=1}^N\), where \(x_i\) denotes the $i$-th material structure and \(P_i \) is the corresponding physical properties. The trained regressor is then used to predict properties \(\hat{P} = (\hat{E}, \hat{I}, \hat{V}_f)\) from a generated structure:
\(
\hat{P} = \mathcal{R}(X).
\)
The prediction accuracy provides a quantitative measure of how well the generated microstructure conforms to the target physical properties embedded in the conditioning prompt.

\noindent\textbf{FiLM-Based Text Embedding Integration:}
To further strengthen conditioning, a FiLM module \citep{perez2018film} modulates intermediate denoising features based on the text embedding. Given noisy input \(x_t\), let \(F \in R^{C \times D \times H \times W}\) be an intermediate feature tensor and \(e \in R^{D}\) the text embedding. FiLM applies an affine transformation:
\begin{equation}
    F' = \gamma \cdot F + \beta,
\end{equation}
where \(\gamma, \beta\) are generated by linear projections of \(e\). This feature-wise modulation allows the model to adaptively optimize microstructure generation according to both semantic descriptions and quantitative physical property guidance.

\subsection{Dual Alignment of 3D Structure and Text}
We propose a dual alignment strategy to enhance the consistency between generated 3D microstructures and their corresponding textual descriptions and physical properties. First, two separate encoders aligns the structure and text embeddings during pre-training. Second, at test time, reward-guided sampling is used to optimize the structure based on the feedback from the learned reward model. These steps jointly enhance the fidelity of both semantics and structure.

\noindent\textbf{Contrastive Text–Structure Alignment:}
Inspired by previous work \citep{radford2021learning, parelli2023clip, chen2024clip}, we employ two separate encoders~\citep{devlin2019bert, tran2018closer} pretrained to align textual and structural representations within a shared embedding space via contrastive learning. Let \( d_i \) denote a textual description of the $i$-th material and \( x_i \in {R}^{D \times H \times W} \) be the corresponding 3D material structure. The text encoder \( f_\theta \) maps \( d_i \) into a text embedding \( z_i^{d} = f_\theta(d_i) \in{R}^D \), while the visual encoder \( g_\phi \) encodes the structure \( x_i \) into a structure embedding \( z_i^x = g_\phi(x_i) \in{R}^D \). 

We adopt a symmetric contrastive alignment objective inspired by cross-modal similarity distribution matching. For each training sample, the similarity logits between modalities are computed as \(
S_{i,j} = \langle z_i^{d}, z_j^x \rangle / \tau,
\) where \( \tau \) is a temperature parameter. To provide soft supervision targets, we further construct intra-modal similarity matrices \(S_{i,j}^{dd}\), \(S_{i,j}^{xx}\) for both modalities and average them to obtain a target similarity distribution. The soft targets are defined as:
\begin{align}
        S_{i,j}^{dd} = \langle z_i^{d}, z_j^{d} \rangle, \quad
    S_{i,j}^{xx} = \langle z_i^x, z_j^x \rangle, \\
    T_{i,j} = \mathrm{softmax}\left( \frac{2(S_{i,j}^{dd} + S_{i,j}^{xx})}{\tau} \right).
\end{align}
Based on these targets, we compute the final alignment loss in a bidirectional manner across modalities:
\begin{align}
    &\mathcal{L}_{forward} = - \sum_{i=1}^N \sum_{j=1}^N T_{i,j} \log \mathrm{softmax}(S_{i,j}), \\
    &\mathcal{L}_{backward} = - \sum_{i=1}^N \sum_{j=1}^N T_{j,i} \log \mathrm{softmax}(S_{j,i}), \\
    &\mathcal{L}_{align} = \frac{1}{2N} \left( \mathcal{L}_{forward} +  \mathcal{L}_{backward} \right).
\end{align}
This loss aligns material descriptions with 3D structures by matching cross-modal similarities to the average of intra-modal similarity patterns.

\noindent\textbf{Test-Time Reward-Guided Alignment:}
To improve the semantic relevance and structural fidelity of generated microstructures, we propose test-time reward-guided alignment. Starting from initial diffusion-based samples, this method iteratively edits local regions using reward feedback. In each round, multiple candidates are sampled, evaluated, and the best edits are retained. Then, a soft resampling step selects the next input from the reward-weighted pool of best structures. 
To balance semantic consistency and structural plausibility, we design two reward models: 
\begin{itemize}
\item \textbf{Contrastive reward:} This reward is computed as the cosine similarity between text and structure embeddings extracted by a visual and text encoders pre-training with CLIP-style text-structure alignment. 
\item \textbf{Discriminative reward:} This reward indicates the score reflecting the authenticity of generated microstructures. We design a discriminator network with 3D convolutional layers, batch normalization, and fully connected layers to distinguish between real and generated structures. This network is trained with binary cross-entropy loss using voxels of both real and generated 3D structures. During testing, it generates a score which represents the fine-grained structural plausibility. 
\end{itemize}

To adapt to the scales of the two rewards, we design a weighted normalization to evaluate the quality of the generated structure. For the $k$-th candidate structure of the $i$-th textual prompt, the final reward score is computed as:
\begin{equation}
    R_{i,k}= \tilde{R}_{i,k}^c + w \cdot \tilde{R}_{i,k}^d,
\end{equation}
where $w$ is the weight hyperparameter, $\tilde{R}_{i,k}^c$ and $\tilde{R}_{i,k}^d$ are the normalized contrastive and discriminative rewards, respectively. Normalization is carried out by subtracting the mean and dividing by the standard deviation within each batch. 




\section{Experiment}

\subsection{Datasets and Evaluation Metrics}
To evaluate the text-guided diffusion model for material structure generation, we conduct comprehensive experiments using the \textbf{Geometries 2000} dataset comprising 2000 text-structure pairs of various types of materials~\citep{zheng2024text}. 

\begin{figure}[t]
    \centering
    \includegraphics[width=0.5\textwidth]{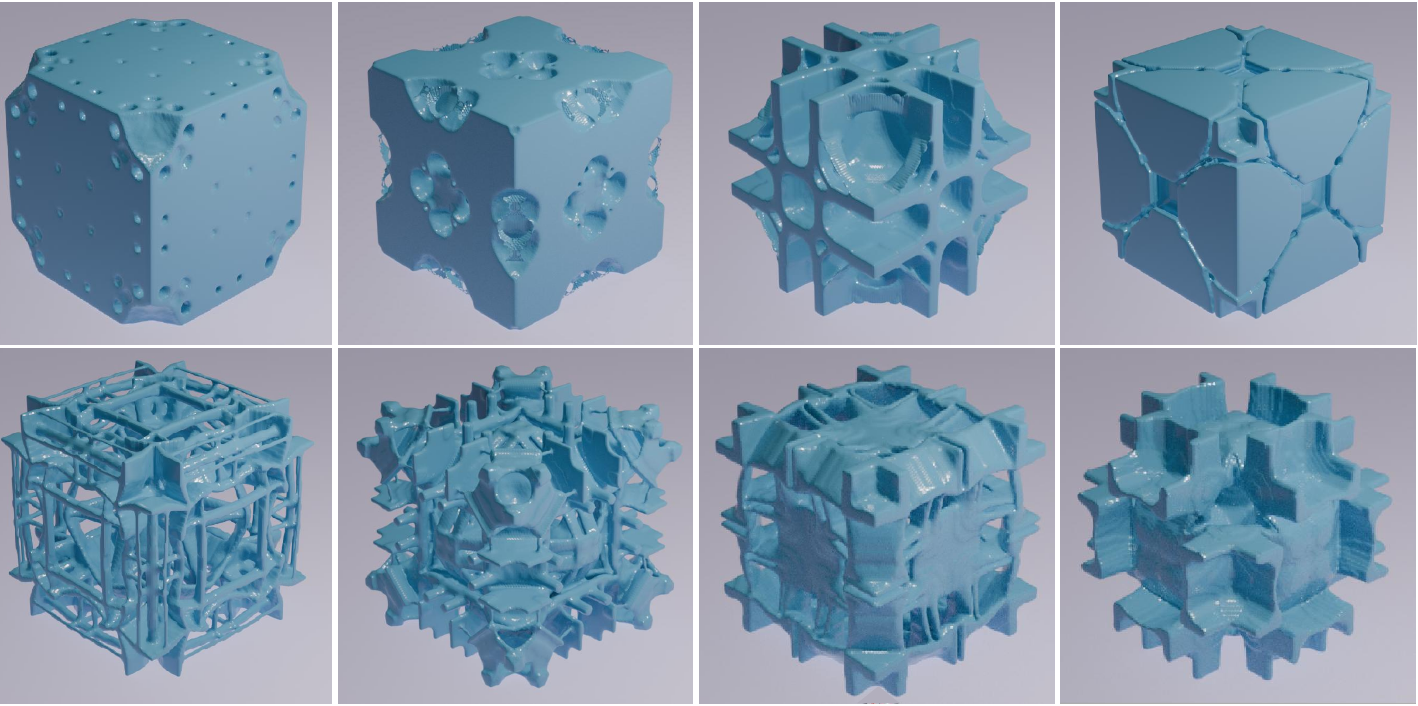}
    \caption{\textbf{Sample 3D structures from GenText-Microstruct:} The dataset comprises metamaterials spanning a broad range of physical properties, distinguishing it from Geometries 2000~\citep{zheng2024text}.}
    \label{fig:new_dataset_vis}
\end{figure}

Due to the limited scale of the Geometries 2000, we construct \textbf{GenText-Microstruct} with textual descriptions, a large text–to-microstructure dataset derived from mechanical metamaterials~\citep{yang2024guided}. Descriptions are first generated by GPT based on conditions including properties, and then manually verified. The dataset contains over 14,000 samples for training and 2,000 for evaluation, covering wide ranges of modulus and Poisson’s ratio. Representative structural examples are illustrated in Figure~\ref{fig:new_dataset_vis}.

We perform a quantitative analysis using four complementary metrics: classification accuracy, Fréchet Inception Distance (FID), CLIP score, and Chamfer Distance (CD). These metrics respectively measure semantic consistency, distributional fidelity, text-to-structure alignment, and geometric accuracy of the generated microstructures. 

\begin{table*}[h]
  \centering
  \setlength{\tabcolsep}{3pt} 
  \caption{\textbf{Quantitative comparisons on the Geometries 2000~\cite{zheng2024text}:} The first four metrics are calculated using 200 random samples from the 2000 datasets, after setting a random seed. For comparison with Txt2Microstruct-Net, the FID score is calculated using 200 prompts that generate 10 prompts with 2000 data. R$^2$-square is used to assess the goodness of fit between the properties of the 2000 true structure and the predicted properties of the generated structure. Our baseline framework first aligns textual and structural semantics, and then performs direct unconstrained diffusion-based generation of 3D structures.}
   \begin{tabular*}{\textwidth}{@{\extracolsep{\fill}} p{13em} c c c c c >{\centering\arraybackslash}p{6em} >{\centering\arraybackslash}p{6em}}
    \toprule
    Method & Accuracy $\uparrow$ & FID $\downarrow$ & CLIP $\uparrow$ & CD $\downarrow$ & R$^2$-square $\uparrow$ & Input & Method \\
    \midrule
    Txt2Microstruct-Net \cite{zheng2024text} & 0.8695 & 72.08 & 0.5599 & 0.0932 & 0.773 0.795 0.771 & Text & VAE \\
    Baseline & 0.8959 & 186.54 & 0.5856 & 0.0694 & 0.849 0.772 0.886 & Text & Diffusion \\
    PropDiff-TMG (ours) & \textbf{0.9100} & \textbf{70.81} & \textbf{0.6936} & \textbf{0.0395} & \textbf{0.961 0.928 0.956} & Text & Diffusion \\
    \bottomrule
  \end{tabular*}
  \vspace{-2mm}
  \label{tab:comparison}
\end{table*}

\begin{figure*}[t]
    \centering
    \includegraphics[width=0.95\textwidth]{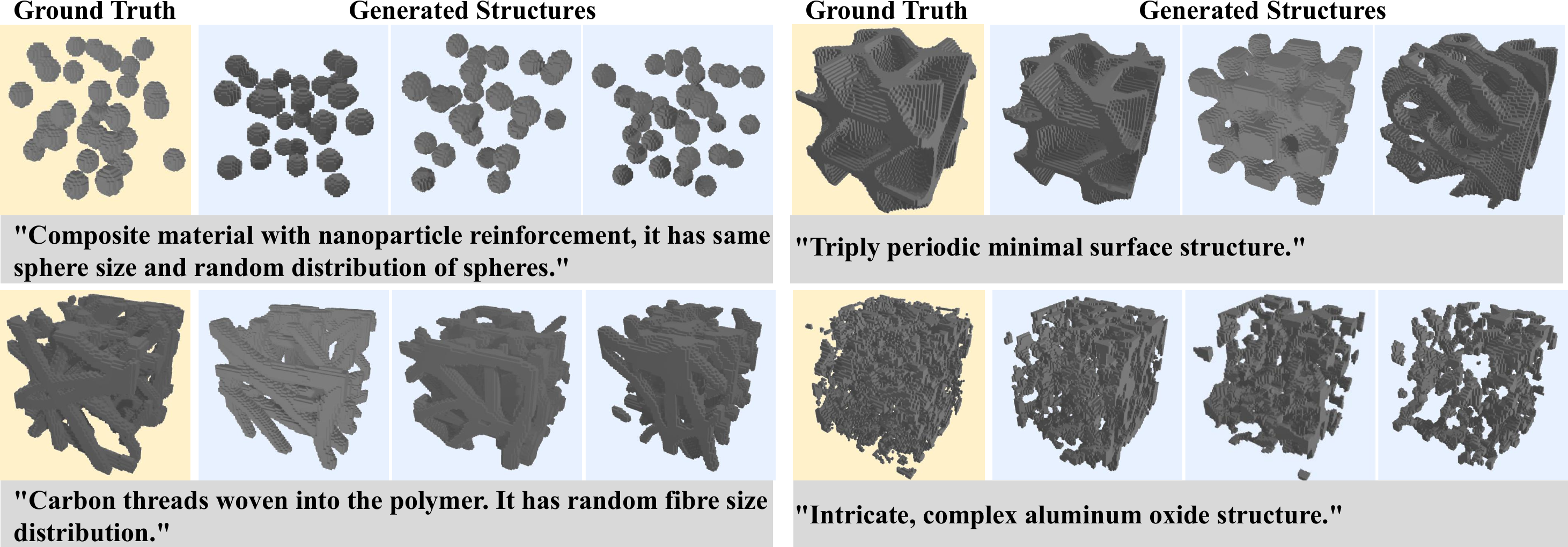}
    \vspace{-0.2cm}
    \caption{\textbf{Qualitative visualizations:} Voxel-based microstructures generated by our model using textual prompts.}
    \label{fig:quality}
    \vspace{-0.1cm}
\end{figure*}


\begin{figure*}[tbp]
    \centering
    \begin{subfigure}[b]{0.32\textwidth}
        \includegraphics[width=\textwidth,height=0.18\textheight,keepaspectratio=false]{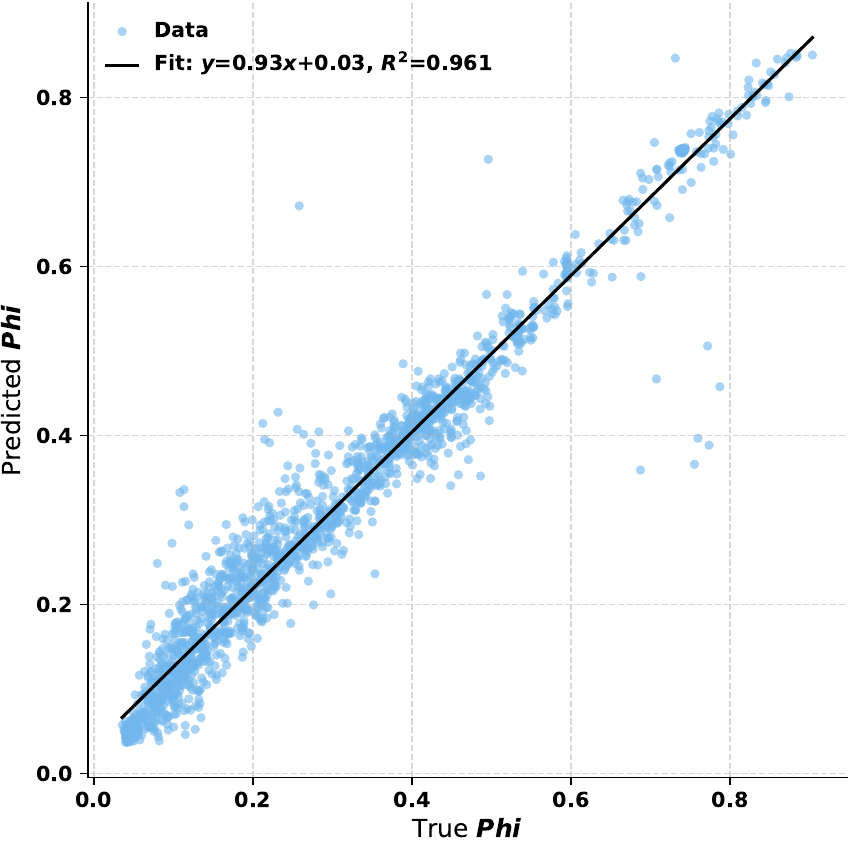}
        \caption{Volume score comparison}
        \label{fig:sub-a}
    \end{subfigure}
    \hfill
    \begin{subfigure}[b]{0.32\textwidth}
        \includegraphics[width=\textwidth,height=0.18\textheight,keepaspectratio=false]{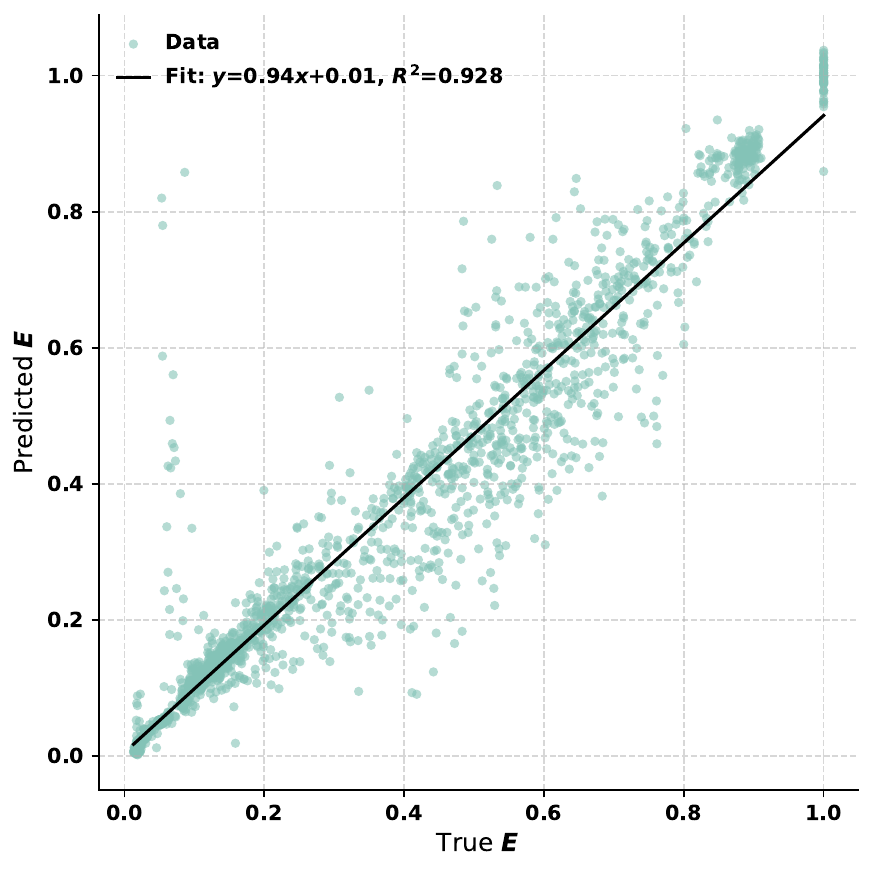}
        \caption{Young's modulus comparison}
        \label{fig:sub-b}
    \end{subfigure}
    \hfill
    \begin{subfigure}[b]{0.32\textwidth}
        \includegraphics[width=\textwidth,height=0.18\textheight,keepaspectratio=false]{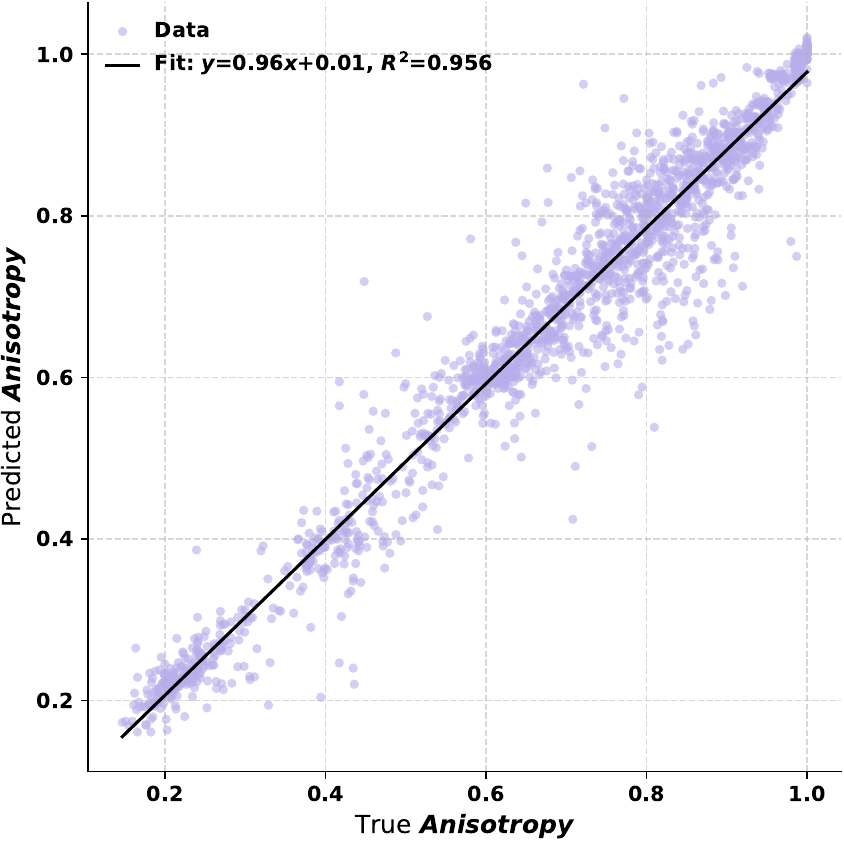}
        \caption{Anisotropy comparison}
        \label{fig:sub-c}
    \end{subfigure}
    \vspace{-0.2cm}
    \caption{\textbf{Comparison of structural physical properties:} Linear regression plots between the true values and the generated physical properties for 2,000 generated microstructures with property conditions. Each subplot compares one property: (a) volume fraction, (b) Young's modulus, and (c) anisotropy score. Higher linearity and lower deviation from the fitted line indicate higher generated accuracy.}
    \label{fig:main}
\end{figure*}

\subsection{Results of Text-to-Microstructure Generation}
We conduct both qualitative and quantitative evaluations to assess the ability of model to generate microstructures under various textual conditions. 

\noindent\textbf{Evaluation on Geometries 2000:} Table~\ref{tab:comparison} shows microstructure generation results on the Geometries 2000.
Our method achieves superior accuracy, suggesting that the generated structures exhibit a stronger correspondence with the intended target labels. Besides, the lower FID indicates that the generated microstructures more closely match the feature distribution of real samples. 

Furthermore, with property-conditioned constraints, the structures exhibit more accurate predicted properties. The higher coefficient of determination and improved linear regression performance, as illustrated in Figure~\ref{fig:main}, demonstrate that our approach is more effective in inverse design that satisfies specified physical constraints. Meanwhile, property error is calculated to verify the mechanical feasibility of the generated structures, as shown in Table~\ref{tab:error}. Additionally, we incorporate the CLIP score to evaluate the quantitative alignment between input conditions and generated structures. The data shows that the generated results are semantically coherent with their descriptions.

\noindent\textbf{Qualitative Analysis:} Rather than relying on design templates, the model autonomously constructs diverse structures guided by natural text. As shown in Figure~\ref{fig:quality}, for a single textual prompt, the model is able to generate multiple outputs with different structures, reflecting the inherent one-to-many nature of text-to-structure mapping. This generation diversity under consistent semantic constraints helps expand the material design space and promotes the discovery of novel or unconventional microstructures. Furthermore, the high consistency between textual and structural semantics allows the model to effectively capture textual semantics and improve manufacturing feasibility. In addition, visual results generated from GPT-based text prompts are provided in the supplementary material.

\begin{table}[h]
  \centering
  \vspace{-3mm}
  \caption{\textbf{Property error results.}}
  \vspace{-2mm}
  \resizebox{0.48\textwidth}{!}{
  \begin{tabular}{lccc}
    \toprule
    Method & Young's modulus $\downarrow$ & Anisotropy $\downarrow$ & Volume fraction $\downarrow$ \\
    \midrule
    Txt2Microstruct-Net & \textbf{0.0118} & 0.0163 & 0.0348 \\
    PropDiff-TMG (ours) & 0.0175 & \textbf{0.0106} & \textbf{0.0103} \\
    \bottomrule
  \end{tabular}}
  \vspace{-4mm}
  \label{tab:error}
\end{table}

\begin{table}[t]
  \centering
  \caption{\textbf{Quantitative comparisons on the GenText-Microstruct:} The baseline is based on properties prompt to generate structures, which follows the strategies used in previous study~\citep{yang2024guided}. Reward-Guided Align denotes the module of Test-Time Reward-Guided Alignment. Since CLIP measures the semantic correspondence between the text description and the generated structure, the baseline performs poorly on this metric.}
  \resizebox{0.48\textwidth}{!}{
  \begin{tabular}{lccc}
    \toprule
    Method & FID $\downarrow$ & CLIP $\uparrow$ & CD $\downarrow$ \\
    \midrule
    Baseline~\citep{yang2024guided} & 84.46 & 0.3281 & 0.0666 \\
    \midrule
    PropDiff-TMG (ours) & \textbf{47.74} & \textbf{0.6463} & \textbf{0.0442} \\
    w/o Property Condition & 52.94 & 0.5210 & 0.0482 \\
    w/o Reward-Guided Align & 49.02  & 0.5164 & 0.0468 \\
    \bottomrule
  \end{tabular}}
  \label{tab:new_dataset}
\end{table}

\noindent\textbf{Evaluation on GenText-Microstruct:} 
Table~\ref{tab:new_dataset} showcases our performance on the GenText-Microstruct dataset. It shows that our method significantly reduces FID and CD scores relative to baseline, reflecting improvements in fidelity and distribution realism, and further demonstrating the robustness and effectiveness of the proposed framework. Furthermore, both property-informed stochastic conditioning and test-time reward-guided alignment improved the performance of each metric, demonstrating the effectiveness and robustness of each module.

\begin{table}[t]
  \centering
  \vspace{-2mm}
  \caption{\textbf{Ablation studies:} Contrastive Align and Reward-Guided Align denote Contrastive Text–Structure Alignment and Test-Time Reward-Guided Alignment, respectively.}
  \vspace{-2mm}
  \resizebox{0.48\textwidth}{!}{
  \begin{tabular}{clccc}
    \toprule
     & Method & FID $\downarrow$ & CLIP $\uparrow$ & CD $\downarrow$ \\
    \midrule
    \textcircled{1} & PropDiff-TMG (ours) & \textbf{70.81} & \textbf{0.6936} & \textbf{0.0395} \\
    \midrule
    \textcircled{2} & w/o Property Condition & 105.35 & 0.6816 & 0.0651 \\
    \textcircled{3} & w/o Contrastive Align & 264.63  & 0.5161 & 0.0579 \\
    \textcircled{4} & w/o Reward-Guided Align & 81.68 & 0.6078 & 0.0412 \\
    \midrule
    \textcircled{5} & w/o Discriminator       & 73.51 & 0.7038 & 0.0396 \\
    \textcircled{6} & w/o Normalization       & 77.52  & \textbf{0.7189} & \textbf{0.0394} \\
    \bottomrule
  \end{tabular}}
  \vspace{-2mm}
  \label{tab:abl_prop}
\end{table}


\begin{figure*}[t]
    \centering
    \includegraphics[width=0.95\textwidth]{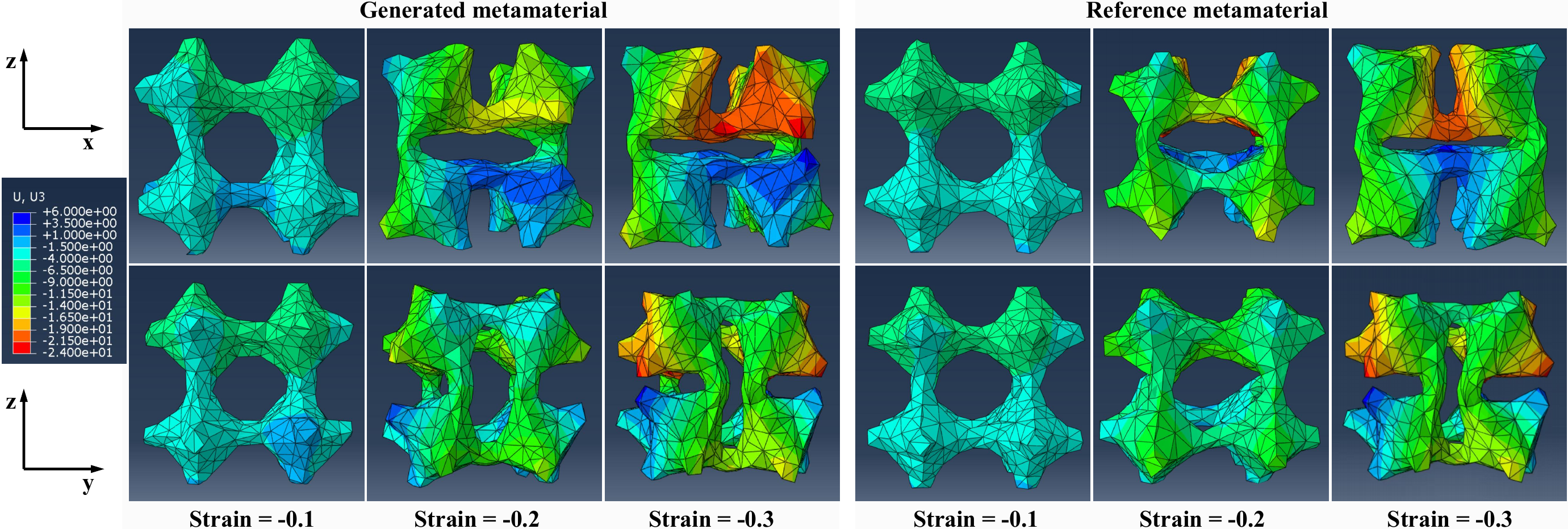}
    \caption{\textbf{Qualitative results of simulations:} Qualitative comparison of mechanical properties of generated metamaterial with those of the reference. Gradual deformed shapes at different compressive strains obtained from finite element simulations.}
    \label{fig:simulation}
\end{figure*}

\begin{figure}[tbp]
    \centering
        \begin{subfigure}[b]{0.235\textwidth} 
            \includegraphics[width=\textwidth]{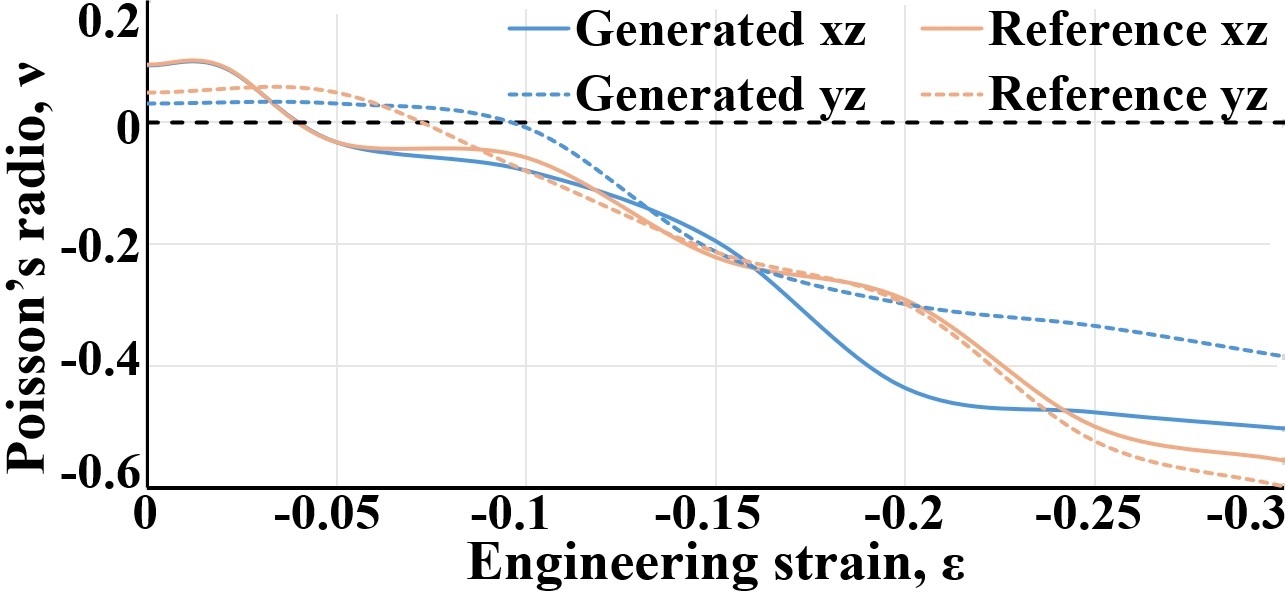}
            \caption{Poisson's-strain curve}
            \label{fig:sub-a}
        \end{subfigure}
        \begin{subfigure}[b]{0.235\textwidth}
            \includegraphics[width=\textwidth]{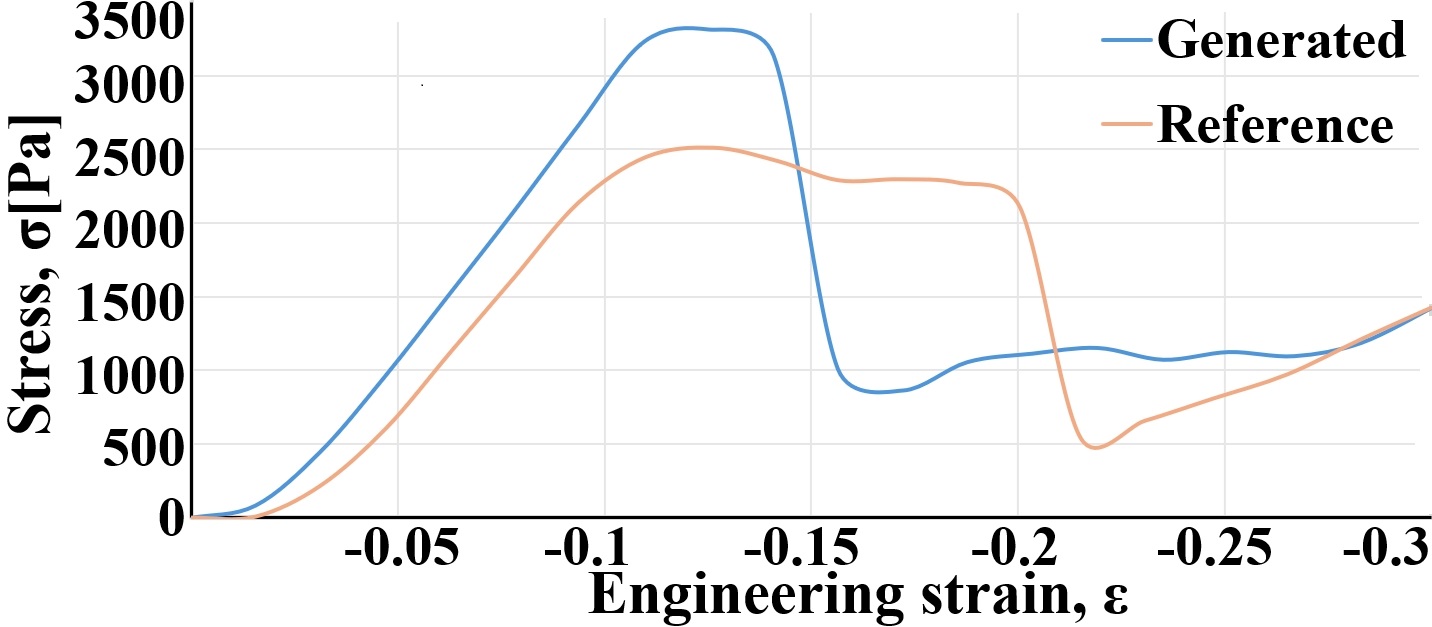}
            \caption{Stress-strain curve}
            \label{fig:sub-b}
        \end{subfigure}
\vspace{-0.5cm}
    \caption{\textbf{Mechanical properties by simulation:} Quantitative comparison of (a) Poisson's ratio strain curve and (b) stress-strain curve between the generated material and the reference material obtained from finite element simulation.}
    \label{fig:simulation2}
\end{figure}

\subsection{Ablation Studies}
To comprehensively evaluate the contributions of different components in our method, we conduct ablation studies on different modules, as shown in Table~\ref{tab:abl_prop}.

\noindent\textbf{Effectiveness of Property-Informed Stochastic Conditioning:}
To evaluate the effect of conditioning on physical properties, we trained the model with additional supervision using property values such as Young’s modulus, volume fraction, and anisotropy. It found that even without providing any physical properties during inference, the model trained with these conditionings produced higher quality structures than the baseline model trained without these conditionings, with the CD score dropping significantly from 0.0651 to 0.0395. This suggests that incorporating physical property information during training can help guide the model to learn more physically meaningful and structurally consistent representations. 

\noindent\textbf{Effectiveness of Contrastive Text-Structure Alignment:}
To evaluate the effectiveness of contrastive text-structure alignment, we directly use the self-conditional diffusion model for training instead of pre-aligning the text and structure. It found that incorporating contrastive text-structure alignment during pre-training significantly improved performance across multiple metrics, with the FID score exhibiting a substantial reduction from 264.63 to 70.81, and the CLIP score significantly increased from 0.5161 to 0.6936. This is primarily due to the incorporation of specialized representations of the material domain during pre-training. This effectively mitigated the semantic bias of general language models in understanding material-related text and enhanced the semantic consistency between structure and description.

\noindent\textbf{Effectiveness of Test-Time Reward-Guided Alignment:}
To evaluate the effectiveness of test-time reward-guided alignment, which refines the generated structure by iteratively updating random regions with candidate regions that have higher rewards. This strategy significantly improves various metrics, demonstrating its ability to guide the model towards the desired semantic and structural goals without additional training. 

In addition, we conduct ablation experiments on each component in this module. When there is only a reward model for text-structure alignment, the CLIP metric is significantly improved to 0.7038. The normalized combined reward model, after adding the discriminator reward model, not only improves the CLIP metric to 0.6936, but also significantly reduces the FID score to 70.81. This shows that the reward model for text-structure alignment acts on semantic consistency, while the discriminator reward model acts on structural rationality.

\subsection{Finite Element Method Simulation}
Finite element method simulations are performed using ABAQUS to investigate the large-deformation behavior of the generated metamaterial structure and a reference metamaterial. The large-deformation response of the auxetic metamaterial is investigated using a geometrically nonlinear static analysis. The material is modeled using a nearly incompressible neo-Hookean hyperelastic model with a Young's modulus set to 0.6615 MPa. To capture buckling-related behavior, a linear eigenvalue buckling analysis is first performed. subsequently, a postbuckling simulation is performed, gradually compressing the structure along the axial direction (z) by a prescribed displacement, terminating the analysis when contact is detected between opposing boundaries. The postbuckling stress-strain response and the evolution of the Poisson's ratio are extracted by tracking the reaction forces and displacements on the loaded surface and the lateral surfaces, respectively. 

Figure \ref{fig:simulation} shows a sequence of the progressively deformed metamaterial generated under four different levels of compressive engineering stress, obtained from the finite element results. Compared with the reference material, the generated metamaterial exhibits favorable negative Poisson ratio properties in both the x- and y-directions, although the deformation process is relatively distorted. 

Figure \ref{fig:simulation2} provides a detailed comparison of the Poisson's ratio-strain and stress-strain curves obtained from simulations. The results show that the Poisson's ratio of the generated structure decreases with increasing compressive strain. Some fluctuations in the curve may be due to the simulation setup, which also affects the stress-strain curve. Although the generated structure may have defects that cause stress to drop, it still exhibits a mechanical trend similar to that of the reference structure during deformation. Experimental results demonstrate that the generative model based on the diffusion model can produce structures that are consistent with the semantic and physical properties targets.

\section{Conclusion}
In this work, we propose PropDiff-TMG, a unified diffusion framework for inverse design of 3D metamaterial microstructures from textual descriptions and physical properties. To optimize the microstructure, we introduce quantitative properties in addition to feature description. Furthermore, a dual alignment strategy is proposed to eliminate cross-domain knowledge differences through text–structure alignment and to strengthen the alignment of semantic and physical structures through a test-time reward-guided alignment. Extensive experiments show that PropDiff-TMG is able to generate diverse, physically plausible, and semantically consistent microstructures, achieving moderate improvement over existing microstructure methods.

\appendix
\section{Test-Time Reward-Guided Alignment}
\begin{figure}[t]
	\centering
	\includegraphics[width=0.5\textwidth]{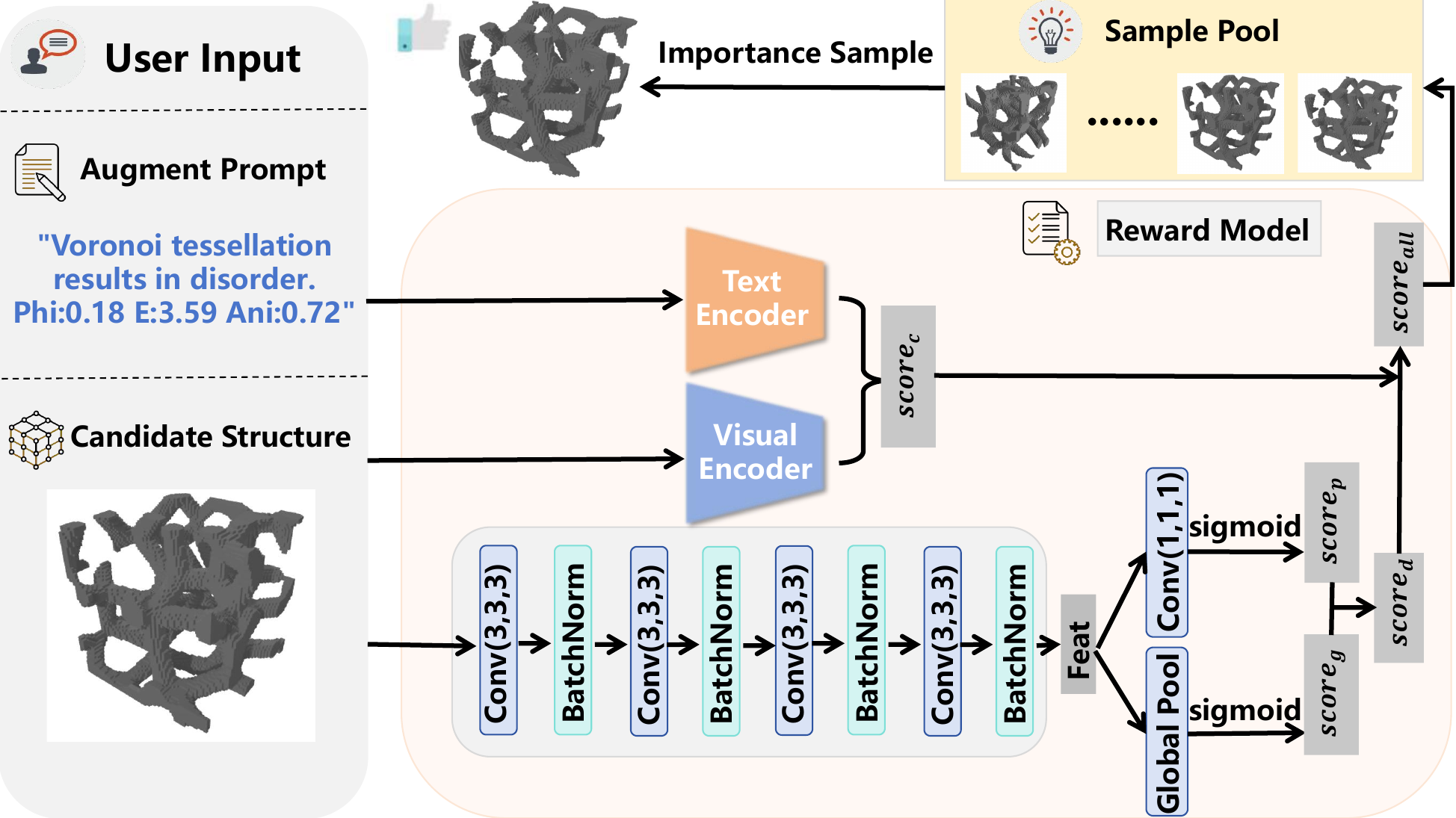}
	\caption{\textbf{Overview of Test-Time Reward-Guided Alignment:} Given an initial material structure and a target textual description, the proposed module performs localized optimization of random regions guided by metric scores derived from two reward models: a contrastive reward and a discriminative reward, thereby progressively refining the 3D material structure toward the target.}
	\label{fig:reward2}
\end{figure}

In the reward model, we employ a pre-trained CLIP-style dual encoder and discriminator to jointly evaluate the semantic alignment and structural plausibility of the generated microstructures, which is illustrated in Figure~\ref{fig:reward2}. 
The discriminator is trained to distinguish high-quality text-aligned structures from generated samples. 
The discriminator's local structure score is denoted as $s_p$, the discriminator's global structure score is denoted as $s_g$, and the overall score $s_f$ is obtained by averaging the patch-level and global probabilities. The weights of CLIP and the discriminator are 1.0 and 0.4, respectively.

Detailed process of test-time reward-guided alignment is shown in Algorithm~\ref{alg:1}. 
The initial sample undergoes \( R = 5 \) iterative optimization steps. In each iteration, \( E = 4 \) local patches of size \( 32 \times 32 \times 32 \) are randomly selected as candidate editing regions. For each patch, \( K = 8 \) candidate structures are generated. These candidates are evaluated using a reward model that weights the semantic similarity score from CLIP and the structural realism score from the discriminator. For each local region, the candidate with the highest reward score replaces the original patch. After each round of optimization, the current best reward is normalized using a softmax function, and the structure from the pool of best resampled samples is used as the starting point for the next round. This iterative process continues until a predetermined number of rounds is reached.


\begin{algorithm}[h]
	\caption{Test-Time Reward-Guided Local Sampling}
	\label{alg:1}
	\begin{algorithmic}[1]
		\State \textbf{Input:} Diffusion model $\mathcal{G}$, prompt $T$, batch size $B$, diffusion steps $S$, guidance scaling factor $w$, edits per round $E$, candidates per edit $K$, total rounds $N$, temperature $\lambda$
		
		\State \textbf{Output:} Optimized structures $X^*$
		
		\State Initialize $X$ from the diffusion model $\mathcal{G}(T, B, S, w)$
		
		\State Set $X^* \leftarrow X$, and compute initial rewards $R^*$
		
		\For{$n=1$ to $N$}
		\State Sample $E$ patch locations  
		\For{$e=1$ to $E$}
		\For{$k=1$ to $K$}
		\State Local edit X with half-step sampling $\hat{X}^{(k)}$  
		\State Replace patch $e$ with a new sample  
		\EndFor
		\State Evaluate the combined rewards of $K$ candidates  
		\State Update $X$ with the highest reward  
		\EndFor
		\State Compute current rewards $R$  
		\If{Reward enhancement}
		\State Update $X^*$ and $R^*$
		\EndIf
		\State Compute probabilities $p = \mathrm{softmax}(\lambda R^*)$  
		\State Sample indices $s \sim \mathrm{Multinomial}(p,B)$  
		\State Set $X \leftarrow X^*[s]$
		\EndFor
		\State \Return $X^*$
	\end{algorithmic}
\end{algorithm}

\section{Evaluation Metrics}
To evaluate the performance of our generated microstructure, we conduct a quantitative analysis using four complementary metrics: classification accuracy, Fréchet Inception Distance (FID), CLIP score, and Chamfer Distance (CD). These metrics respectively assess the semantic consistency, distributional fidelity, text-to-structure alignment, and geometric accuracy of the generated microstructures. Furthermore, we used the R-squared coefficient to evaluate the correlation between the predicted properties of the generated structure and the predicted properties of the real structure.
\subsection{Classification Accuracy}
It can evaluate the ability of the model to generate microstructures that are recognizable and distinguishable across predefined categories. To perform this evaluation, a 3D convolutional neural network classifier is trained using 2,000 labeled microstructures distributed over 20 semantic classes. The high accuracy achieved by the classifier on the generated samples indicates strong semantic alignment between the structures and the textual prompt.

\subsection{FID Score}
To evaluate the generation quality of 3D microstructure, we adopt the Fréchet Inception Distance (FID) metric by extracting feature embeddings from the penultimate layer of a pre-trained 3D classifier. We compute the FID score between 2,000 real and 2,000 generated microstructures using the standard formulation:
\begin{equation}
	{FID=||\mu_x-\mu_y||}^2+Tr(C_x+C_y-2{(C_xC_y)}^{1/2})
\end{equation}
where \( x \) and \( y \) represent the feature vectors of real and generated structures, respectively, \( \mu_x \) and \( \mu_y \) are the mean vectors, \( C_x \) and \( C_y \) are the covariance matrices, and \( Tr \) denotes the matrix trace. A lower FID indicates that the distribution of generated structures is closer to that of real samples, suggesting higher generation fidelity.
\subsection{CLIP Score}
It can measure the degree of semantic alignment between the input text and the corresponding microstructure generated. A pretrained dual encoder from the first stage of training is employed to embed both the text description and the generated microstructure into a shared representation space. Semantic alignment is subsequently assessed by computing the cosine similarity between their corresponding embeddings. A high similarity score indicates that the model effectively captures and preserves the semantic content of the input prompt in the generated output.

\subsection{CD Score}
It is computed between point clouds sampled from the generated and ground-truth material structures, serving as a measure of geometric similarity. To compute the CD score, the material structures, both generated and ground-truth, are first converted into surface point clouds, typically by sampling the coordinates of boundary voxels or using isosurface extraction methods. Let \(P = \{ p_i \}_{i=1}^N \) and \(Q = \{ q_j \}_{j=1}^M \) represent the two point clouds sampled from the generated and ground-truth structures, respectively. The CD score is written as:
\begin{equation}
	{CD} = \frac{1}{|P|} \sum_{p \in P} \min_{q \in Q} \|p - q\|_2^2 + \frac{1}{|Q|} \sum_{q \in Q} \min_{p \in P} \|q - p\|_2^2
\end{equation}
which ensures that each point in one set is close to at least one point in the other. The first term evaluates how much of the generated structure is relevant to the target geometry, while the second term captures how well the generated structure covers the target. The lower CD score corresponds to higher shape fidelity and more accurate reconstructions.

\subsection{R-Squared Coefficient}
It can evaluate the ability of the model to generate microstructures that closely match the target physical properties. For this evaluation, we train a solver with the same convolutional architecture as the classifier using 2000 properties-labeled microstructures. A higher R-squared coefficient, approaching one, indicates stronger physical consistency in the generated structures.

\section{Implementation Details}
During the pre-training phase, PropDiff-TMG trains a property-informed dual encoder using a contrastive text-structure pre-training approach to align text-structure embeddings. The dataset contains \(N = 2000\) microstructures, with 80\% used for training and 20\% for validation. Each text description is accompanied by the physical properties of its corresponding microstructure, including volume fraction, effective Young’s modulus, and isotropy. To improve generalization, the physical property is randomly masked during training. A pre-trained 3D-ResNet model~\citep{tran2018closer} serves as the structure encoder, and a pre-trained BERT model~\citep{devlin2019bert} serves as the text encoder. The two encoders are jointly trained to optimize the similarity matrices of structure and text embeddings. 
The model trains within a self-conditional diffusion framework, processing input microstructures at a resolution of \(64 \times 64 \times 64\) voxels. 
Training is performed using the Adam optimizer with a learning rate of \(1\times10^{-4}\), a batch size of 64, and a temperature parameter of 0.1. 
Attention mechanisms operate at spatial resolutions of 4 and 8, utilizing 4 attention heads and a dropout rate of 0.1. Training incorporates an exponential moving average (EMA) with a decay rate of 0.999 to stabilize parameter updates.
Training runs for 2000 epochs on a single NVIDIA GeForce RTX 4090 GPU. The inference time is approximately 5 seconds per sample with 100 diffusion steps on a single RTX 4090 GPU, while the optimized testing-time inference takes 110.4 seconds.
\begin{figure*}[t]
	\centering
	\includegraphics[width=1\textwidth]{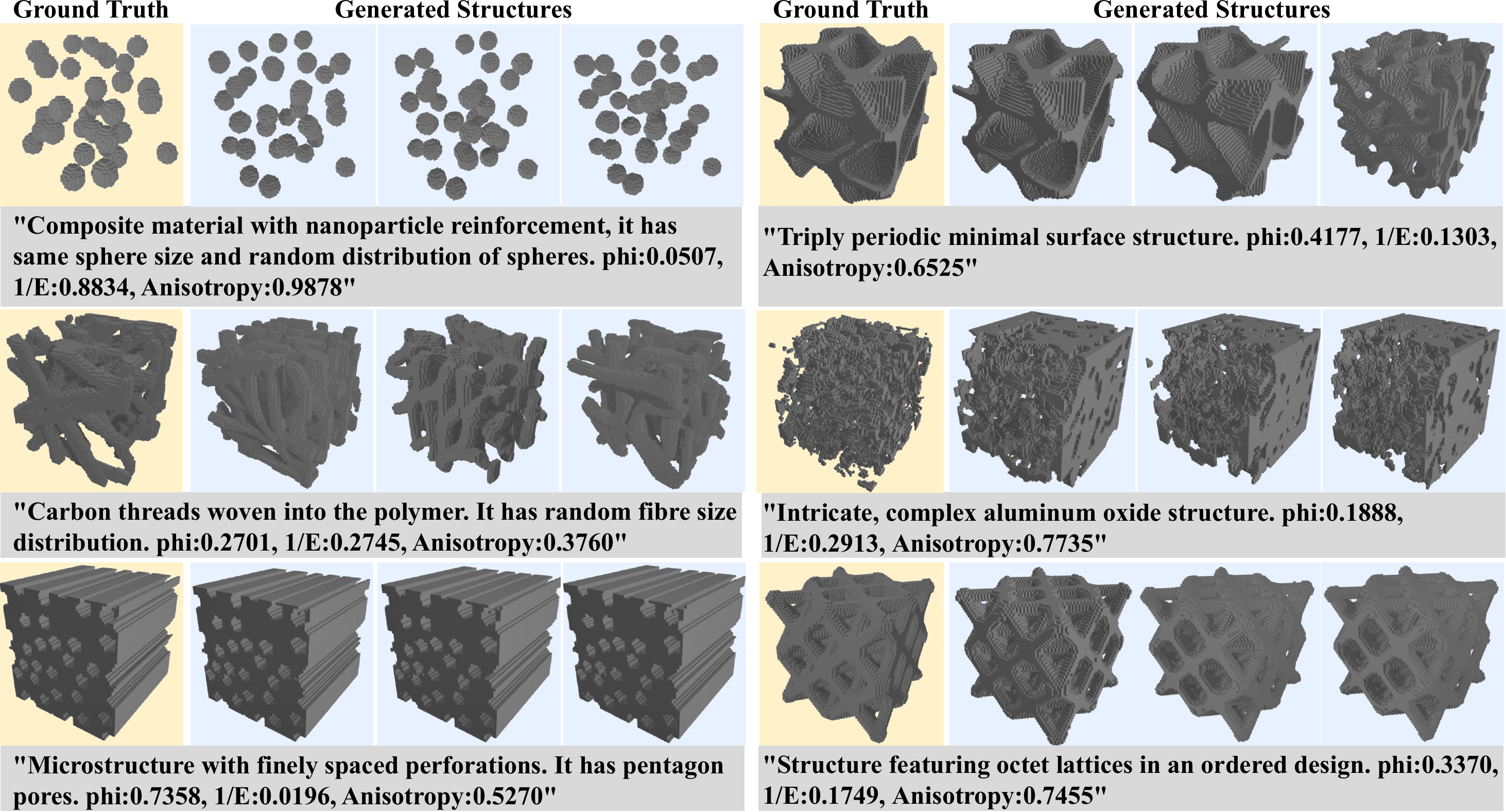}
	\caption{\textbf{More qualitative visualizations:} Voxel-based microstructures generated by our model using textual prompts with properties.}
	\label{fig:quality3}
\end{figure*}

\begin{figure}[h]
	\centering
	\includegraphics[width=0.5\textwidth]{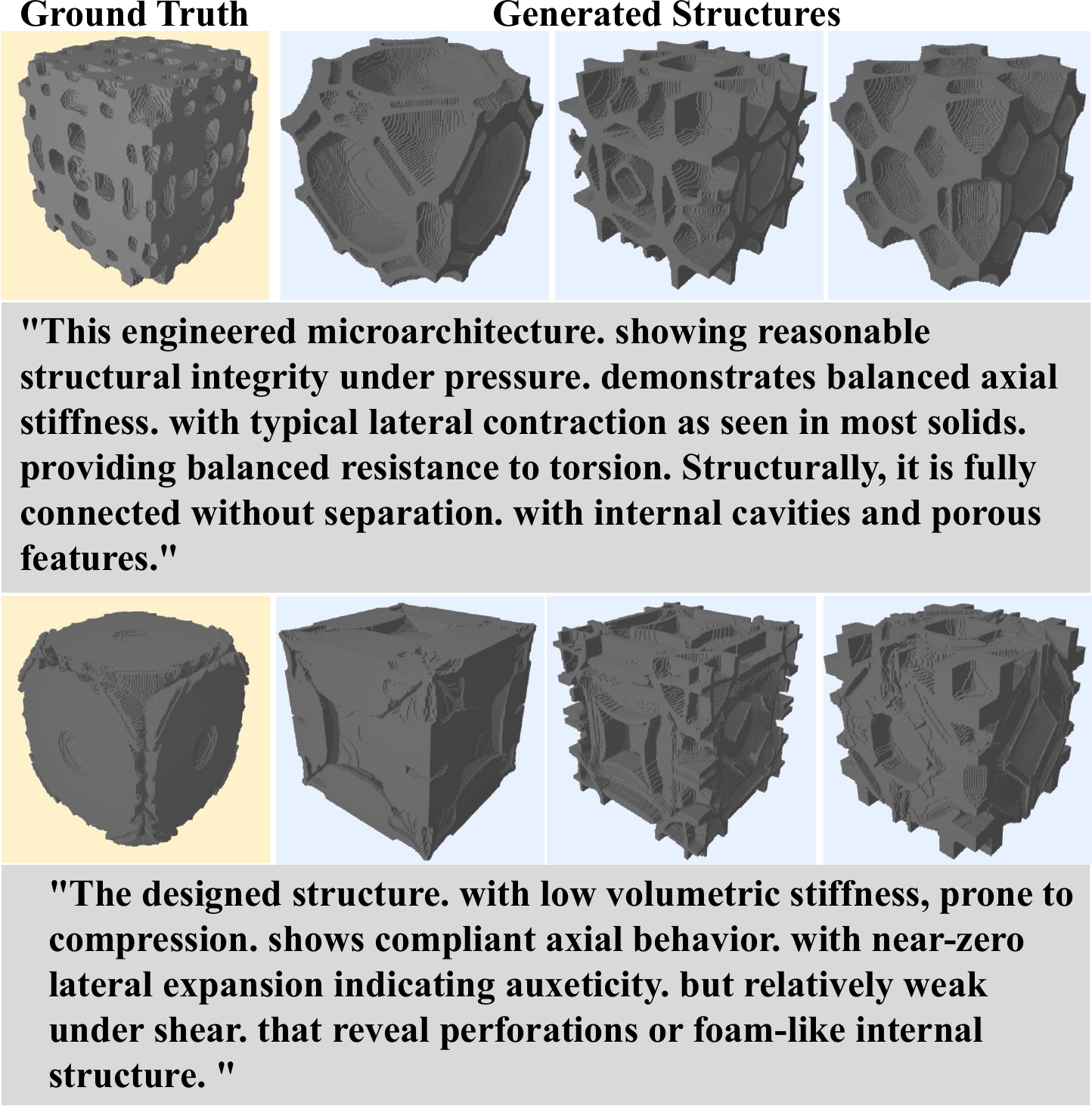}
	\caption{\textbf{Qualitative visualizations of GenText-Microstruture without physical properties:} Voxel-based mechanical metamaterial structures are generated in response to text prompts produced by a rule-based text generator.}
	\label{fig:newdata}
\end{figure}

\begin{figure}[h]
	\centering
	\includegraphics[width=0.5\textwidth]{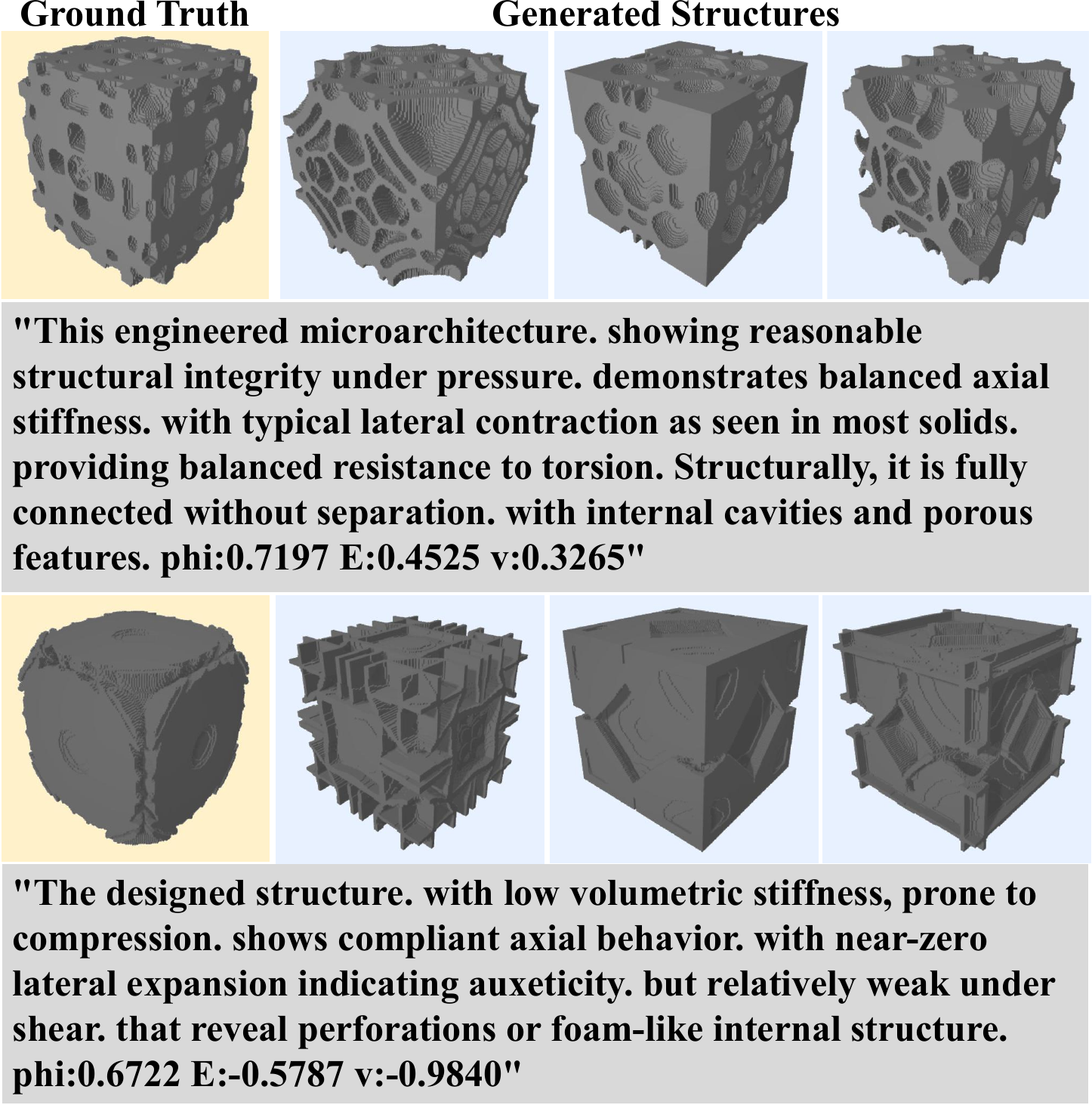}
	\caption{\textbf{Qualitative visualizations of GenText-Microstruture with physical properties:} Voxel-based mechanical metamaterial structures are generated based on attributed text prompts generated by a rule-based text generator.}
	\label{fig:newdata_prop}
\end{figure}

\section{Experiment}
\subsection{Dataset Construction}
GenText-Microstructure is a text-structure pair dataset based on automatically generated microstructure physical descriptions from 3D voxels~\citep{yang2024guided} as prompts. Over 14,000 metamaterial structures spanning a broad spectrum of modulus and Poisson’s ratio are utilized for training, while another 2,000 randomly generated structures are used for evaluation. For each binary voxel, we extract geometric features such as connectivity, Euler number, surface area to volume ratio, symmetry, and internal porosity. In addition, the corresponding stiffness tensor provides mechanical properties, including \( C_{11} \), \( C_{12} \), \( C_{44} \), and the derived bulk modulus \( V \). All data are normalized using a pre-trained scaler. Subsequently, GPT is employed to generate descriptive text conditioned on the properties and features, and a rule-based text generator converts the quantitative properties into a coherent, scientifically styled natural-language description, covering volumetric stiffness, axial stiffness, Poisson's ratio, shear strength, and structural topology.

\subsection{Visualizations and Simulations}
\noindent\textbf{Qualitative Analysis on Geometries 2000:} 
Figure~\ref{fig:quality3} shows the visualization of qualitative results generated based on properties-driven text. The introduction of property constraints makes the generated results more targeted while still meeting the target physical properties, but this also reduces structural diversity.

\noindent\textbf{Qualitative Analysis on GenText-Microstructure:} 
To evaluate the model capacity to interpret and translate structured textual descriptions into diverse material geometries, we conducted qualitative visualization experiments using the GenText-Microstructure. As shown in Figure~\ref{fig:newdata}, the visualization results demonstrate that the model effectively captures the underlying structural semantics while exhibiting diverse results. As shown in Figure~\ref{fig:newdata_prop}, the generated structures after adding properties maintain the overall form while exhibiting local diversity. This indicates that the model not only learns deterministic text-structure mappings but can also generalize to generate different structures under the same text semantics.

\begin{figure*}[tbp]
	\centering
	\includegraphics[width=1\textwidth]{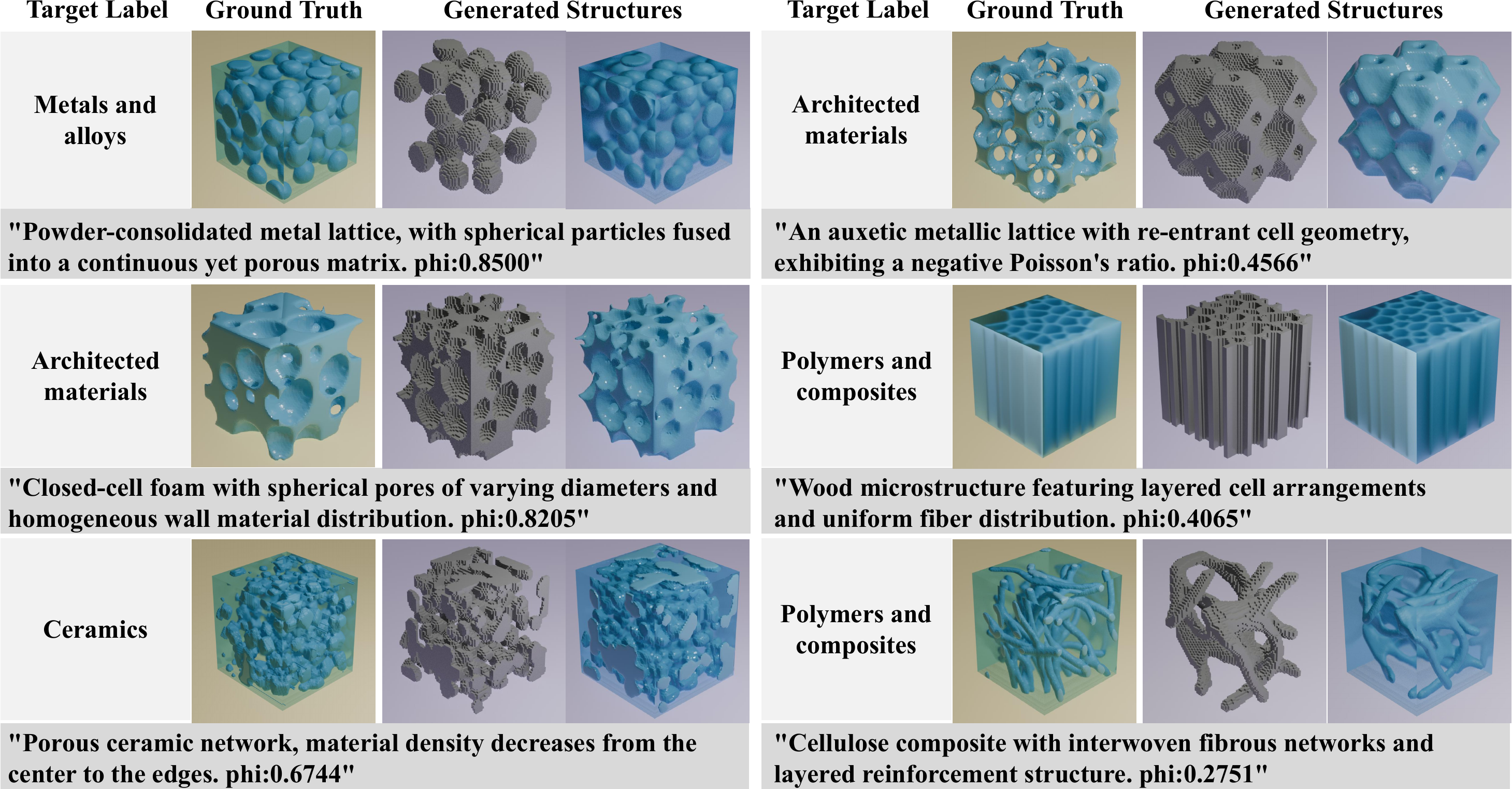}
	\caption{\textbf{Qualitative visualizations of GPT-generated textual descriptions:} 3D structure generation was conducted based on GPT-generated textual descriptions across different material categories. The figure presents representative target structures, the corresponding GPT-generated text prompts, and the resulting generated 3D structures.}
	\label{fig:quality_gpt}
\end{figure*}

\noindent\textbf{Visualization of Simulation:} 
Simulation results are presented in the videos included in the supplementary materials. 
The compression process of the resulting metamaterial structure is visually recorded via video contour plots extracted from a series of nonlinear finite element simulations. These contour plots illustrate the gradual evolution of deformation as the axial compressive displacement is incrementally applied along the \(z\)-axis, starting from 0 to 0.3. The video begins with the undeformed configuration and proceeds through successive loading stages. Throughout the simulation, the displacement field distribution is color-mapped to emphasize regions of high deformation gradients. The progression demonstrates that the generated structure consistently exhibits auxetic behavior, as indicated by lateral compression in the \(x\)- and \(y\)-directions under axial compression. Compared to the reference metamaterial, the generated structure maintains a pronounced negative Poisson’s ratio effect across a wide range of compressive strains. Furthermore, the contour plots reveal localized stress concentrations and deformation distortions. Overall, the results confirm that the generated structures maintain the desired mechanical behavior.

\subsection{Visualization under Diverse Prompts}
To evaluate model generalization and text–structure alignment beyond the training distribution, we conduct a qualitative visualization experiment using newly generated and diverse text prompts created by GPT. The Geometries 2000~\citep{zheng2024text} includes four types of materials: metals and alloys, polymers and composites, ceramics, building materials, and metamaterials. For each material category, we select a representative target structure and textual descriptions generated by GPT describing its morphology and spatial distribution. The model then generates 3D structures based on each textual prompt. As shown in Figure~\ref{fig:quality_gpt}, the generated samples demonstrate that the model can accurately capture structural semantics from various textual descriptions. This confirms that the model can generalize beyond memorized training samples and align textual semantics with structure generation in an interpretable manner.

\subsection{Ablation Experiment}
To comprehensively evaluate the contribution and generalization ability of different components in our method, we conduct ablation studies on different modules on GenText-Microstruture, as shown in Table~\ref{tab:abl_prop}.

\noindent\textbf{Effectiveness of Property-Informed Stochastic Conditioning:}
To verify the effectiveness of property-informed stochastic conditioning, we conduct additional ablation experiments on our constructed text-structure dataset. As shown in Table~\ref{tab:abl_prop}, the results show that adding physical properties further improves the model across all metrics, indicating higher consistency in semantics among the generated structures. Furthermore, this module effectively reduces ambiguity in structure generation, enabling the model to more accurately align expected physical features while maintaining diversity.

\begin{table}[tbp]
	\centering
	\caption{\textbf{Ablation studies:} Contrastive Align and Reward-Guided Align denote Contrastive Text–Structure Alignment and Test-Time Reward-Guided Alignment, respectively. In evaluating the FID metric, we select the visual encoder from Contrastive Text-Structure Alignment as the feature extractor.}
	\resizebox{0.48\textwidth}{!}{
		\begin{tabular}{clccc}
			\toprule
			& Method & FID $\downarrow$ & CLIP $\uparrow$ & CD $\downarrow$ \\
			\midrule
			\textcircled{1} & Ours & \textbf{47.74} & \textbf{0.6463} & \textbf{0.0442} \\
			\midrule
			\textcircled{2} & w/o Property condition & 52.94 & 0.5210 & 0.0482 \\
			\textcircled{3} & w/o Reward-Guided Align & 49.02  & 0.5164 & 0.0468 \\
			\midrule
			\textcircled{5} & w/o Discriminator       & 51.89 & 0.6396 & \textbf{0.0404} \\
			\textcircled{6} & w/o Normalization       & 51.94 & 0.6396 & 0.0454 \\
			\bottomrule
	\end{tabular}}
	\label{tab:abl_prop}
\end{table}

\noindent\textbf{Effectiveness of Test-Time Reward-Guided Alignment:}
To validate the effectiveness and generalization ability of the test-time reward-guided alignment module, we conduct additional ablation experiments on our constructed text-structure dataset. The text prompts on this dataset are generated using regularized templates, systematically describing different structural features, and differed from the original dataset in text-structure distribution. As shown in Table~\ref{tab:abl_prop}, the results show that the proposed reward-guided alignment still significantly improves performance across multiple metrics, further confirming that this module can guide the model to generate structures that are more functionally and physically consistent without additional training.

Furthermore, additional ablation experiments are conducted on each component of the module to validate the effectiveness of the discriminator and the normalized combined reward function. As shown in Table~\ref{tab:abl_prop}, the results demonstrate that incorporating the discriminator slightly increases the CD score but improves FID and CLIP, indicating enhanced diversity and semantic fidelity instead of target memorization. Meanwhile, the normalized combined reward function balances the contributions of the two reward metrics, avoiding a single factor dominating the optimization process, thereby improving all metrics. This further demonstrates the effectiveness of these two modules.

\section*{Acknowledgments}
This work was supported by National Science Foundation of China (52441503, 62302093), Jiangsu Province Natural Science Fund (BK20230833), the CIPS-SMP-Zhipu Large Model Fund, the Fundamental Research Funds for the Central Universities (2242025K30024), the Open Research Fund of the State Key Laboratory of Multimodal Artificial Intelligence Systems (E5SP060116) and the Big Data Computing Center of Southeast University.

{
    \small
    \bibliographystyle{ieeenat}
    \bibliography{main}
}

\end{document}